\pdfoutput=1
\documentclass[11pt]{article}

\usepackage[]{EMNLP2023}

\usepackage{times}
\usepackage{latexsym}
\usepackage{multirow}
\usepackage{graphicx}
\usepackage{enumitem}
\usepackage{amsmath}
\usepackage{amssymb}
\usepackage{latexsym}
\usepackage[inkscapelatex=false]{svg}
\usepackage{url}

\usepackage[linesnumbered,ruled,vlined]{algorithm2e} 
\usepackage{subfigure}
\usepackage{makecell}
\usepackage{color}
\usepackage{booktabs}
\usepackage{hyperref}

\usepackage{listings}
\usepackage{color}

\definecolor{dkgreen}{rgb}{0,0.6,0}
\definecolor{gray}{rgb}{0.5,0.5,0.5}
\definecolor{mauve}{rgb}{0.58,0,0.82}

\usepackage[all]{nowidow}
\usepackage[T1]{fontenc}

\usepackage[utf8]{inputenc}

\usepackage{microtype}

\usepackage{inconsolata}

\title{ACL Anthology Helper: A Tool to Retrieve and Manage \\ Literature from ACL Anthology}

\author{Chen Tang\textsuperscript{1}, Frank Guerin\textsuperscript{1} and Chenghua Lin\textsuperscript{2}\footnotemark[1] \\
\textsuperscript{1}Department of Computer Science, The University of Surrey, UK \\
\textsuperscript{2}Department of Computer Science, The University of Manchester, UK \\
\texttt{\{chen.tang,f.guerin\}@surrey.ac.uk}\\
\texttt{chenghua.lin@manchester.ac.uk}
} 
\begin{document}
\maketitle

%  for authors' footnotes
\renewcommand{\thefootnote}{\fnsymbol{footnote}} 
\footnotetext[1]{Corresponding author.} 
\renewcommand*{\thefootnote}{\arabic{footnote}}

\begin{abstract}
The ACL Anthology is an online repository that serves as a comprehensive collection of publications in the field of natural language processing (NLP) and computational linguistics (CL). This paper presents a tool called ``ACL Anthology Helper''. It automates the process of parsing and downloading papers along with their meta-information, which are then stored in a local MySQL database. This allows for efficient management of the local papers using a wide range of operations, including "where," "group," "order," and more. By providing over 20 operations, this tool significantly enhances the retrieval of literature based on specific conditions. Notably, this tool has been successfully utilised in writing a survey paper~\cite{tang2022recent}. By introducing the ACL Anthology Helper, we aim to enhance researchers' ability to effectively access and organise literature from the ACL Anthology. This tool offers a convenient solution for researchers seeking to explore the ACL Anthology's vast collection of publications while allowing for more targeted and efficient literature retrieval.
\end{abstract}

% =============================== Section 1 ==================================
\section{Introduction}
The ACL Anthology serves as a valuable resource for researchers in the fields of NLP and CL, providing access to a diverse collection of academic literature from reputable venues such as ACL, EMNLP, NAACL, and COLING. It offers researchers the opportunity to keep abreast of the latest advancements, explore foundational work, and discover relevant studies in their areas of interest. However, the ACL Anthology lacks advanced functionalities, such as keyword-based retrieval and filtering options based on publication time, authors, and other criteria. To address these limitations, we present the ACL Anthology Helper, a Python-based software tool designed to facilitate the local download and management of literature from the ACL Anthology.

The ACL Anthology Helper is developed under the MIT License and can be found at the following GitHub repository \url{https://github.com/tangg555/acl-anthology-helper}. This tool offers the following key functionalities:

\begin{itemize}[noitemsep,nolistsep,leftmargin=*]
\item Downloading literature from the ACL Anthology website to a local MySQL database, allowing users to specify publication venues and time spans.
\item Supporting original MySQL operations as well as additional chain operations facilitated by ABuilder\footnote{This is implemented by another github repository, ABuilder (\url{https://github.com/lizhenggan/ABuilder})}, enabling users to effectively manage and retrieve downloaded papers using a sequence of Python-like functions.
\item Wrapping each paper's information into an object, facilitating operations such as Union, Intersection, Complement, and rule-based filtering on groups of papers.
\item Providing structured and statistical information on retrieved papers.
\end{itemize}

The primary objective of this paper is to introduce the architecture of the ACL Anthology Helper and demonstrate how researchers can leverage its functionalities to enhance their literature management and retrieval processes.

% ================================================================
\section{Data Structures}
\begin{figure}[t]
\centering
\includegraphics[width=0.99\linewidth]{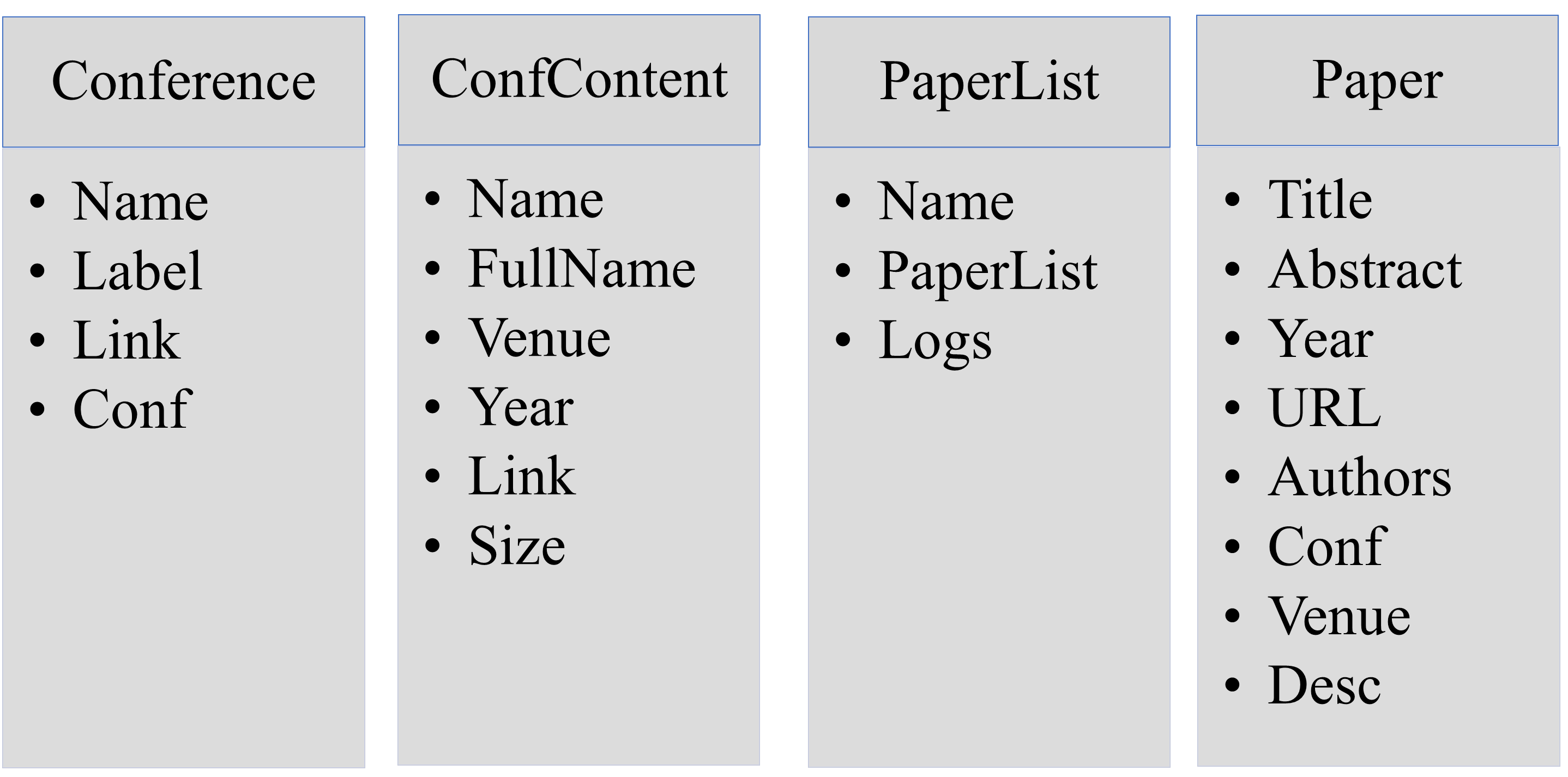}
\caption{The data structure of ACL Anthology Helper. Desc is the abbr. of description. Logs record the information during crawling the data.}
\label{fig:intro}
\end{figure}

Based on the categorization provided by ACL Anthology, as presented in Appendix \ref{apx:homepage}, we have manually devised the data structures depicted in Figure \ref{fig:intro}. These data structures serve as essential components in our iterative crawling process of retrieving literature from the website. ACL Anthology encompasses top-level categories, namely ACL events and Non-ACL events, which comprise venues along with their respective hosting years. Notably, aside from conferences, ACL Anthology incorporates various publication formats, such as journals (TACL), tutorials, and so on. For the convenience of implementation, we consider all of these formats as conferences, as the majority of events within ACL Anthology correspond to conferences and share a similar web structure during the crawling process. 

The basic categories within ACL Anthology are presented as the events, which serve as entries in the conference proceedings. The \textit{Conference} objects, representing each event entry, are initially recorded as the indices of the data. Considering the time-consuming and unnecessary nature of downloading all data from ACL Anthology, our tool allows for partial storage of literature by selecting a specific range of conferences. The \textit{Paper} objects act as the data units recorded for literature retrieval. The \textit{ConContent} object, an extended version of \textit{Conference}, exclusively provides essential information for subsequent web crawling, without being stored locally. Similarly, \textit{PaperList} is not actual data, but a group of objects managing multiple papers to support operations such as Union, Intersection, and Complement.

% ----------- fig -----------
% ----------- end of fig -----------

% ================================================================
\section{Crawling Online Literature}
% ----------- fig -----------
\begin{figure}[t]
\centering
\includegraphics[width=0.7\linewidth]{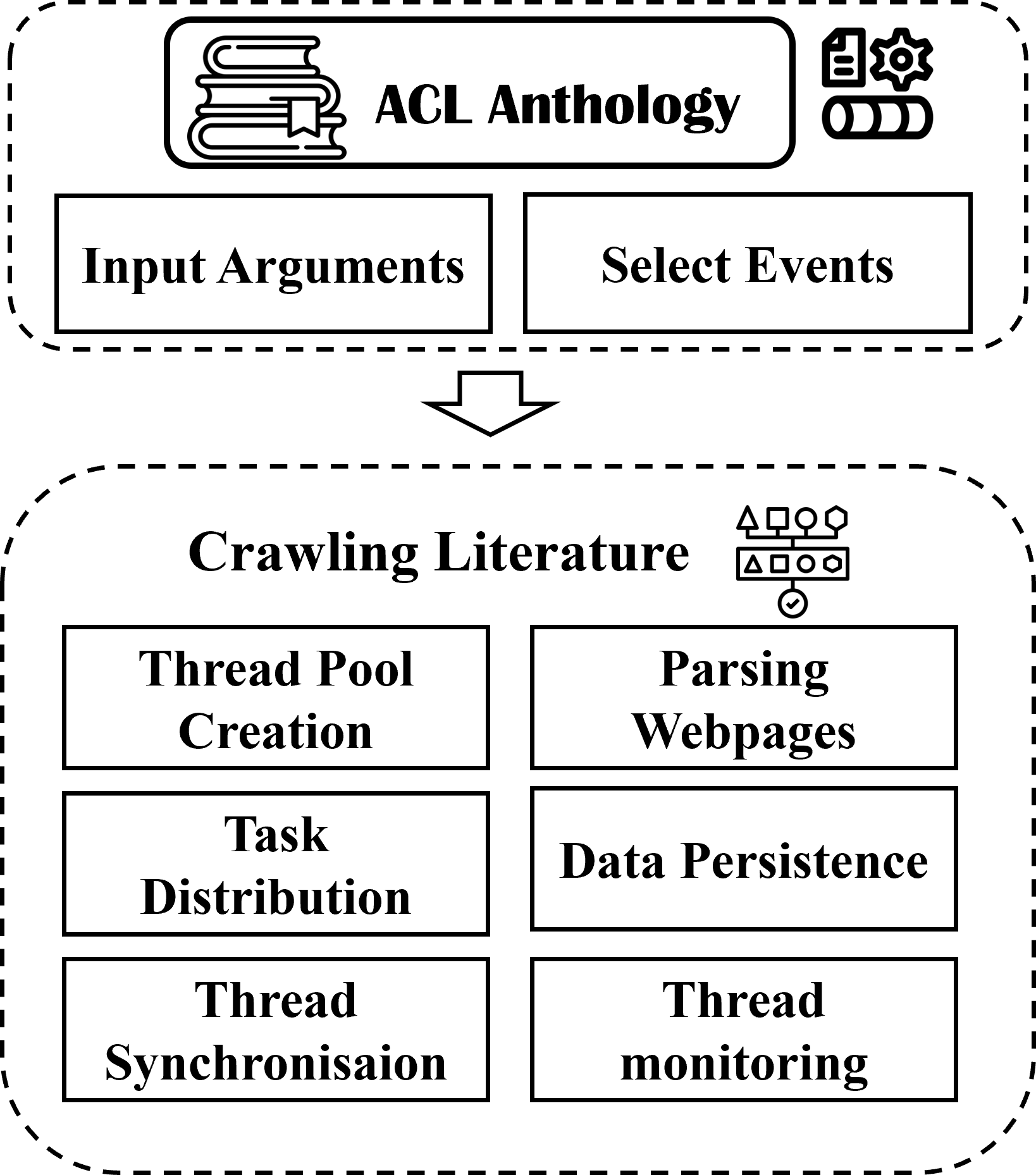}
\caption{The illustration of the online literature crawling process.}
\label{fig:crawling}
\end{figure}
% ----------- end of fig -----------

This section introduces the modules related to crawling online literature for the anthology as shown in \autoref{fig:crawling}. 

\subsection{Webpage Parsing}
The vast collection of literature housed in the ACL Anthology is all publicly accessible. Therefore, utilising a web crawling tool, we can directly retrieve the resources available on this website. In our study, we have implemented an anthology-oriented web crawler  with the widely-used Python package "BeautifulSoup"\footnote{\url{https://github.com/wention/BeautifulSoup4}}. This package is specifically designed for extracting data from HTML and XML files. It provides a convenient and intuitive means of navigating, searching, and modifying the parsed tree by working alongside a parser.

The majority of the literature entries featured in the ACL Anthology consist of static results and do not involve dynamic rendering, encryption, or authorization requirements. As we parse the webpages, the semi-structured data representing the literature is transformed into Python objects. These objects are then persistently stored in local the MySQL database. The parsing process is guided by heuristic rules, facilitating the extraction of relevant information from the webpages.

\subsection{Multi-threaded Execution}
In order to optimise the efficiency and speed of data retrieval from the website of the ACL Anthology, the multi-threaded execution is implemented within the web crawling process. Multi-threading enables the simultaneous execution of multiple threads, allowing for parallel processing and enhanced performance. There are mainly 6 modules for this multi-threaded literature downloading indicated in \autoref{fig:crawling}. By employing multi-threaded execution, we were able to significantly accelerate the crawling process, enabling efficient retrieval of literature data from the ACL Anthology website. 

\textbf{Thread Pool Creation}: It creates a pool of threads to manage the concurrent execution of tasks. The size of the thread pool was determined based on factors such as the available computational resources and the desired degree of parallelism. \textbf{Task Distribution}: The crawling process involved extracting literature resources from multiple webpages within the ACL Anthology website. We divided these webpages into smaller tasks and assigned them to different threads within the thread pool. This ensured that multiple threads could work simultaneously on different parts of the website, enabling efficient data retrieval. \textbf{Thread Synchronisation}: To ensure thread safety and prevent data inconsistencies, we implemented appropriate thread synchronization mechanisms. For instance, we employed locks or semaphores to control access to shared resources, such as databases or data structures. This prevented conflicts or race conditions that could arise from simultaneous access by multiple threads. \textbf{Parsing Webpages}: Within each thread, we utilized the web crawling tool, along with the Python package "BeautifulSoup," to parse the HTML or XML files of the webpages. We extracted the relevant literature data using predefined parsing rules. This step involved navigating the parse tree, searching for specific elements, and extracting the desired information. \textbf{Data Persistence}: As the data was extracted, each thread independently persisted the extracted information into the local databases. Proper synchronization mechanisms were implemented to ensure that data was stored accurately and coherently, even in the presence of concurrent access by multiple threads. \textbf{Thread Monitoring}: Throughout the execution, we monitored the progress and completion of each thread. This allowed us to keep track of the crawling process and identify any potential issues or errors that may have occurred during execution.

% ================================================================
\section{The Database}
% ----------- fig -----------
\begin{figure}[t]
\centering
\includegraphics[width=0.85\linewidth]{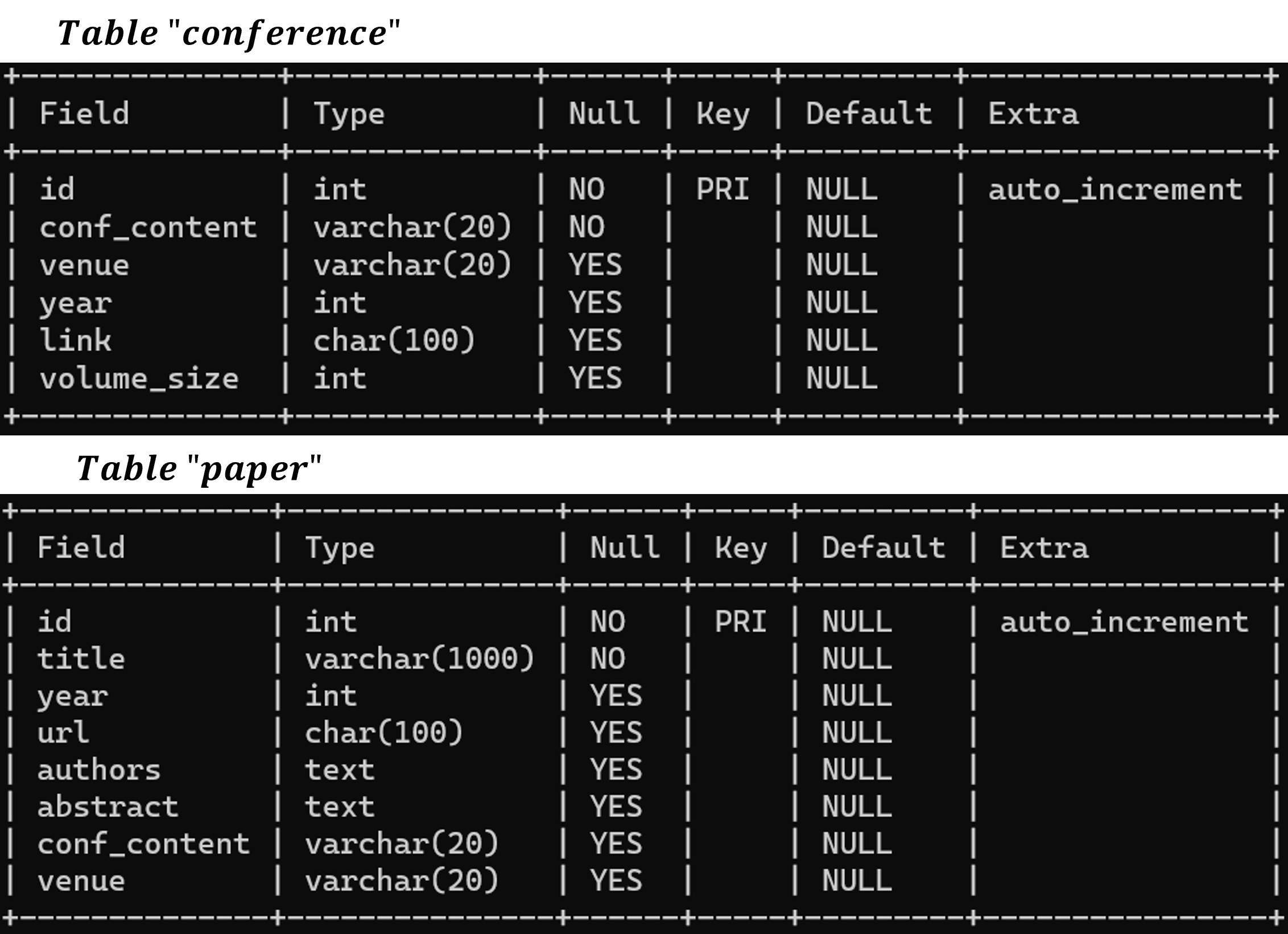}
\caption{The two tables in the ``aclanthology'' database.}
\label{fig:tables}
\end{figure}
% ----------- end of fig -----------

The utilisation of the MySQL~\footnote{\url{https://www.mysql.com/cn/}} database is an essential dependency for our tool, and it is imperative to independently install this database. Following the installation, we configured an initial database named "aclanthology" within MySQL. Subsequently, this configuration was manually populated within the tool's settings files, ensuring proper connectivity. During the execution of the literature downloading code, our tool autonomously generated tables within the "aclanthology" database. These tables, as illustrated in \autoref{fig:tables}, encompassed the representation of conferences and papers. Each table was structured to accommodate specific data attributes. The structured data obtained from the crawling process was then stored within the MySQL database based on the predefined data structures delineated by the two aforementioned tables. 
This systematic storage approach ensured the organization and accessibility of the crawled literature data. By adhering to this database configuration and data storage methodology, we established a robust foundation for the storage and management of the acquired literature data.

% ================================================================
\section{The Retriever}
In addition to the standard SQL commands supported by the aclanthology database, the ABuilder package was employed to introduce an additional set of commands for enhanced data retrieval and manipulation. ABuilder, an open-source Python package available at https://github.com/lizhenggan/ABuilder, was specifically designed to streamline the process of accessing and querying databases. It offers a comprehensive range of functionalities that simplify interactions with databases, providing an intuitive and efficient approach to construct and execute SQL queries tailored to researchers' specific data subset requirements.

To utilise ABuilder, we established a connection to the \textit{aclanthology} database using the connector libraries provided by the package. This connection facilitated seamless access to the database, enabling the retrieval of the desired data. The following code example, depicted in Listing 1, demonstrates the use of ABuilder for constructing complex SQL queries:

\begin{lstlisting}[caption = The code example using ABuilder.]
data = ABuilder().table('paper') \
    .where({"year": ["in", years_limit]}) \
    .where({"venue": ["in", conf_contents]}).query()
papers = MySQLTools.list_to_papers(data)
\end{lstlisting} 

As shown in Listing 1, ABuilder enabled us to construct complex SQL queries by leveraging its query-building capabilities. In this particular example, we retrieved papers that met specific conditions regarding years and conferences. By incorporating various parameters such as conditions, aggregations, and joins, ABuilder facilitated the retrieval of precise data subsets from the aclanthology database. Moreover, ABuilder provided flexibility in handling the retrieved data by offering convenient methods to transform it into suitable formats, such as pandas DataFrames. This feature facilitated subsequent analysis and manipulation of the data.

By harnessing the capabilities of the ABuilder package, we achieved efficient and effective data retrieval from the aclanthology database. Its user-friendly interface and powerful query-building features simplified the retrieval process, enhancing our ability to extract the required data for our research analysis.

Overall, the integration of ABuilder proved to be instrumental in facilitating advanced data retrieval and manipulation from the aclanthology database, ultimately contributing to the successful execution of our research endeavors.

% ================================================================
\section{Software License}
Our tool is released under the permissive MIT License, which imposes minimal restrictions on usage. In alignment with this licensing approach, the project dependencies, including the well-established ABuilder package, are also licensed under the MIT License. Users are granted the freedom to utilize and customize our tool according to their specific requirements. It is worth noting that if users choose to create derivative versions or forks of the tool, we strongly recommend that they also maintain the MIT License to preserve the open nature of the software and foster collaboration within the community.

% =============================== Section 5 ==================================
\section{Discussion}
The primary objective of this software is to enhance the convenience of literature reading. When conducting research in a specific area, the use of general search engines like Google or the online search function of ACL Anthology often returns a large volume of literature, making it challenging to identify high-quality and closely related papers. In such cases, it becomes necessary to apply more stringent and selective conditions to filter the papers effectively. Unlike the original searching service provided by ACL Anthology, which aims to deliver a broader range of results (often in the thousands) with a general prompt, our tool aims to provide a more precise and personalized searching service.

For instance, when exploring recent advancements (spanning from 2021 to 2023) in event-triggered story generation, specifically in the most influential venues such as the top three NLP conferences (ACL, EMNLP, and NAACL), our tool can significantly narrow down the results. Using the ACL browser, the search displays 2040 results. However, with our tool, we can limit the results to just four papers:

\begin{itemize}[noitemsep,nolistsep,leftmargin=*]
\item EtriCA: Event-Triggered Context-Aware Story Generation Augmented by Cross Attention, 2022~\cite{tang-etal-2022-etrica}
\item Persona-Guided Planning for Controlling the Protagonist’s Persona in Story Generation, 2022~\cite{zhang-etal-2022-persona}
\item Long Text Generation by Modeling Sentence-Level and Discourse-Level Coherence, 2021~\cite{guan-etal-2021-long}
\item OpenMEVA: A Benchmark for Evaluating Open-ended Story Generation Metrics, 2021~\cite{guan-etal-2021-openmeva}
\end{itemize}

By leveraging our tool, researchers can obtain a concise selection of relevant papers, enabling them to focus their literature review and analysis on the most significant and pertinent works.

% =============================== Section 6 ==================================
\section{Conclusion}
In this paper, we have presented the ACL Anthology Helper, a software tool aimed at improving the accessibility and organization of literature from the ACL Anthology. By automating the process of parsing and downloading papers, along with their meta-information, this tool enables researchers to efficiently manage their local collection of papers using a variety of operations. The ACL Anthology Helper addresses the limitations of the ACL Anthology by providing advanced functionalities such as keyword-based retrieval, filtering by publication time, authors, and other criteria. Researchers can now retrieve literature based on specific conditions, enhancing their ability to find high-quality and closely related papers in their research area.

% =============================== Acknowledgments ==================================
\section*{Acknowledgements}
Chen Tang is supported by the China Scholarship Council (CSC) for his doctoral study (File No.202006120039).

\bibliography{custom}
\bibliographystyle{acl_natbib}

\appendix

\section{Appendices}
\label{sec:appendix}

\subsection{The homepage of ACL Anthology}
\label{apx:homepage}

\begin{figure}[t]
\centering
\includegraphics[width=0.99\linewidth]{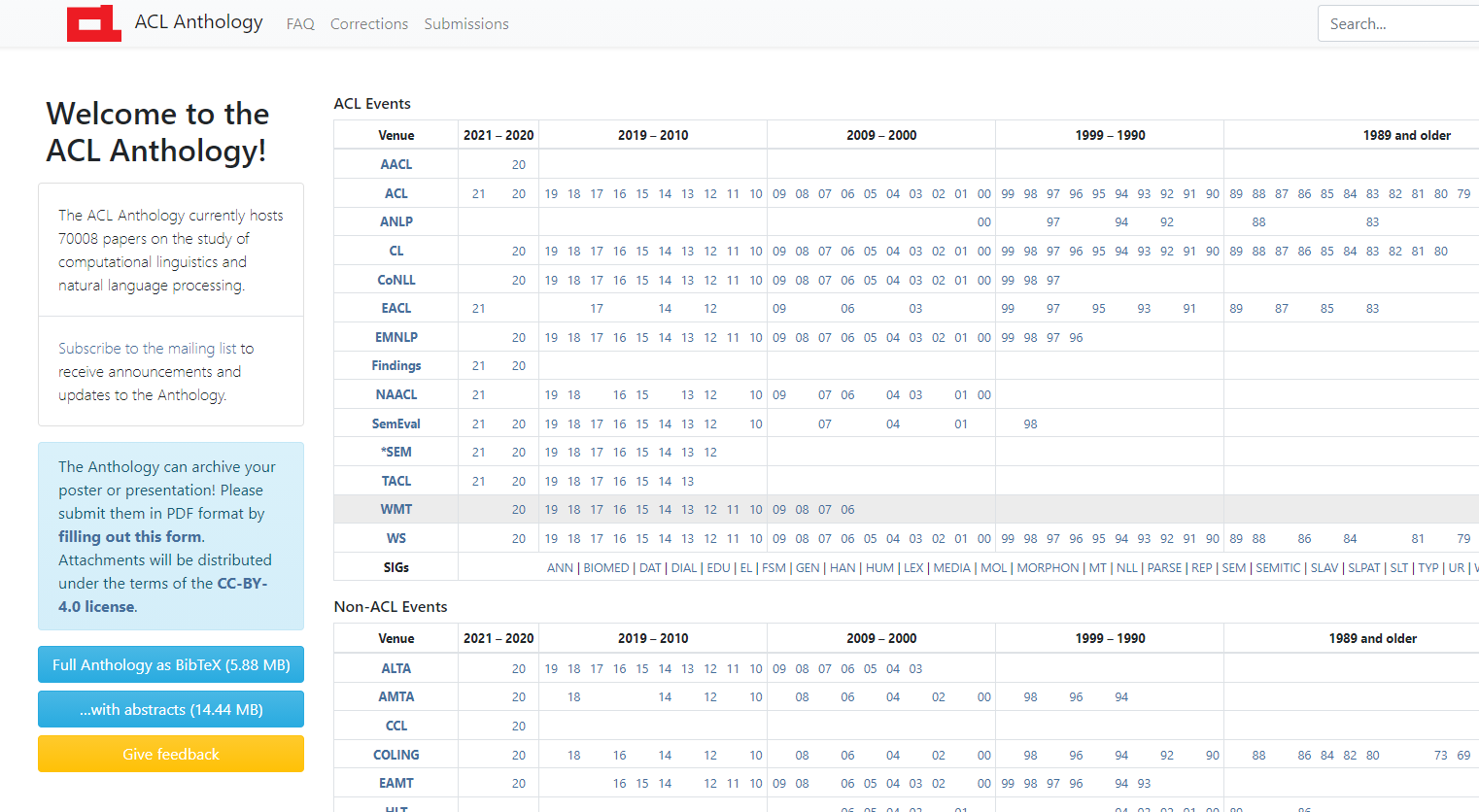}
\caption{The Homepage of ACL Anthology.}
\label{fig:homepage}
\end{figure}

\autoref{fig:homepage} illustrates a screenshot captured from the homepage of the ACL Anthology website.

\subsection{Our other works}
Our research group has actively contributed to the field of Natural Language Generation (NLG) through various scholarly endeavors. We present a comprehensive categorization of our contributions as follows: Story Generation~\cite{tang-etal-2022-ngep,huang-etal-2022-improving,tang-etal-2022-etrica}, Dialogue Generation~\cite{tang2023terminology,tang-etal-2023-enhancing,zhanga2023cadge,yang2023improving}, data-to-text~\cite{yang2023effective}, text summarisation~\cite{tang2023improving,goldsack2023enhancing} and tongue twister generation~\cite{loakman-etal-2023-twistlist}. We believe these works will aid you in navigating our contributions in the field of NLG.

% ----------- fig -----------
% ----------- end of fig -----------

\end{document}